\documentclass[conference]{IEEEtran}
\IEEEoverridecommandlockouts
\usepackage{cite}
\usepackage{amsmath,amssymb,amsfonts}
\usepackage{algorithmic}
\usepackage{graphicx}
\usepackage{textcomp}
\usepackage[dvipsnames]{xcolor}
\def\BibTeX{{\rm B\kern-.05em{\sc i\kern-.025em b}\kern-.08em
    T\kern-.1667em\lower.7ex\hbox{E}\kern-.125emX}}

\usepackage{booktabs}

\usepackage[xindy,style=long,toc=true,nonumberlist,acronym,acronymlists={abbreviations},hyperfirst=true]{glossaries}
\usepackage{algorithm} 
\usepackage{mathtools} 
\mathtoolsset{showonlyrefs=true}
\usepackage{gensymb}
\usepackage{multirow}
\usepackage[caption=false,font=footnotesize]{subfig}
\usepackage{comment}
\usepackage{siunitx}

\newcommand{\etal}{\text{\textit{et al.\@ }}}
\newcommand{\etals}{\text{\textit{et al.}'s\@ }}

\usepackage[frozencache,cachedir=.]{minted}
\definecolor{LightGray}{gray}{0.9}

\newacronym{ML}{ML}{machine learning}
\newacronym{DL}{DL}{deep learning}
\newacronym{IoU}{IoU}{intersection over union}
\newacronym{SVM}{SVM}{support vector machine}
\newacronym{XAI}{XAI}{explainable AI}
\newacronym{DT}{DT}{decision tree}
\newacronym{AI}{AI}{artificial intelligence}
\newacronym{CNN}{CNN}{convolutional neural network}
\newacronym{DC}{DefChar}{defect characteristic}
\newacronym{RGB}{RGB}{red, green and blue}
\newacronym{HSV}{HSV}{hue, saturation, and brightness}
\newacronym{TPR}{TPR}{true positive rate}
\newacronym{TNR}{TNR}{true negative rate}
\newacronym{RCNN}{R-CNN}{region-based convolutional neural network}
\newacronym{IEMRCNN}{IE-MRCNN}{image-enhanced mask R-CNN}
\newacronym{SHAP}{SHAP}{shapley additive explanations}
\newacronym{XAI360}{XAI360}{explainable AI 360}
\newacronym{FM}{FM}{Forest Monkey}
\newacronym{MN}{CCMN}{COVID-CT-mask-net}
\newacronym{LFCN}{LFCN}{lightweight fully convolutional network}
\newacronym{RUN}{ResUNet++}{deep residual U-Net++}

\begin{document}
\title{ForestMonkey: Toolkit for Reasoning with AI-based Defect Detection and Classification Models}

\author{\IEEEauthorblockN{Jiajun Zhang, Georgina Cosma\IEEEauthorrefmark{1}}
\IEEEauthorblockA{\textit{Dept of Computer Science, School of Science} \\
\textit{Loughborough University, UK}\\
Email: j.zhang8@lboro.ac.uk; g.cosma@lboro.ac.uk\IEEEauthorrefmark{1}\\
\IEEEauthorrefmark{1} Corresponding author}
\and
\IEEEauthorblockN{Sarah Bugby}
\IEEEauthorblockA{\textit{Dept of Physics, School of Science} \\
\textit{Loughborough University, UK}\\
Email: s.bugby@lboro.ac.uk}
\and
\IEEEauthorblockN{Jason Watkins}
\IEEEauthorblockA{\textit{Railston \& Co. Ltd.}\\ 
\textit{Nottingham, UK}\\
Email: jason@railstons.com}
\thanks{This research is funded through joint funding by the School of Science at Loughborough University with industrial support from Railston \& Co Ltd.}
}

\maketitle

\begin{abstract}
\textit{\Gls{AI}} reasoning and \textit{\gls{XAI}} tasks have gained popularity recently, enabling users to explain the predictions or decision processes of AI models. This paper introduces \textit{\gls{FM}}, a toolkit designed to reason the outputs of any \gls{AI}-based defect detection and/or classification model with data explainability. Implemented as a Python package, \gls{FM} takes input in the form of dataset folder paths (including original images, ground truth labels, and predicted labels) and provides a set of charts and a text file to illustrate the reasoning results and suggest possible improvements. The \gls{FM} toolkit consists of processes such as feature extraction from predictions to reasoning targets, feature extraction from images to \textit{defect characteristics}, and a decision tree-based \gls{AI}-Reasoner. Additionally, this paper investigates the time performance of the \gls{FM} toolkit when applied to four \gls{AI} models with different datasets. Lastly, a tutorial is provided to guide users in performing reasoning tasks using the \gls{FM} toolkit.
\end{abstract}

\glsreset{AI}
\glsreset{XAI}
\glsreset{FM}
\glsreset{DC}

\section{Introduction}
\label{sec:introduction}
Object detection and classification tasks using \textit{\gls{AI}} are popular in industry applications and beyond. The outputs of \gls{AI} models need to be explained to foster trust in the results during decision making. \textit{\Gls{XAI}} techniques offer solutions for explaining and reasoning the outcomes of \gls{AI} models, leading to the development of various \gls{XAI} toolkits. One such toolkit is XAI360 \cite{xai3602019}, published by Arya \etal of the \textit{International Business Machines Corporation (IBM)} in 2020, which integrates 14 diverse \gls{XAI} methods.

In 2023, Ali \etal \cite{xai_survey} conducted a survey on a set of \gls{XAI} techniques and proposed four \gls{XAI} taxonomies: scoop-based, model-based, complexity-based, and methodology-based, along with their corresponding use cases. Scoop-based \gls{XAI} techniques, also known as data explainability methods \cite{de1,de4,de5},
analyse the features extracted from the data to establish the relationship between the input and output of the \gls{AI} model. Model-based \gls{XAI} techniques, such as model explainability methods \cite{ted,me1,me3}, 
break down the \gls{AI} model into steps and provide explanations for each step. Complexity-based \gls{XAI} techniques, such as intrinsic interpretable models (e.g.\ decision trees, Bayesian models, linear models, and k-nearest neighbours), offer explanations with varying levels of detail based on the model's complexity. Methodology-based \gls{XAI} techniques, referred to as post-hoc explainability methods \cite{shap,lime,gce}, 
interpret the \gls{AI} models by analysing the model's backpropagation routes or perturbation signals. Zhang \etal \cite{zhangair} developed an \gls{AI} reasoning framework that enables reasoning capabilities for the outputs of an \gls{AI}-based detection and/or classification model. The framework incorporates a proposed feature extraction method known as \textit{\glspl{DC}} and an ensemble \textit{\gls{DT}}. 

In this paper, a toolkit named \textit{\gls{FM}} is introduced, which implements Zhang \etals \cite{zhangair} framework using the Python programming language. The \gls{FM} toolkit provides reasoning for predictions made by an \gls{AI}-based detection and/or classification model. The \gls{FM} toolkit takes as input images, ground truth labels, and predicted labels from the \gls{AI} model. Subsequently, the toolkit converts the predicted labels into reasoning targets and transforms the images into a \glspl{DC} matrix. Then, the toolkit generates the \gls{AI} reasoning results, which include a set of charts and text-based improvement suggestions. By examining the reasoning results, users can gain a deeper understanding of their dataset and make improvements to their \gls{AI} model.

In this paper, Section~\ref{sec:overview} provides an overview of the structure of the \gls{FM} toolkit. Section~\ref{sec:impl} explains the implementation details of the toolkit. Section~\ref{sec:perf} applies the toolkit to four different \gls{AI}-based models using diverse datasets and discusses the performance in terms of execution time. Finally, Section~\ref{sec:tuto} presents a tutorial on using the toolkit and provides an explanation of the output generated by the toolkit. The code for the \gls{FM} toolkit and tutorial is provided in https://github.com/edgetrier/AI-Reasoner.

\section{Toolkit Overview}
\label{sec:overview}

\begin{figure*}[!t]
    \centering
    \includegraphics[width=0.80\linewidth]{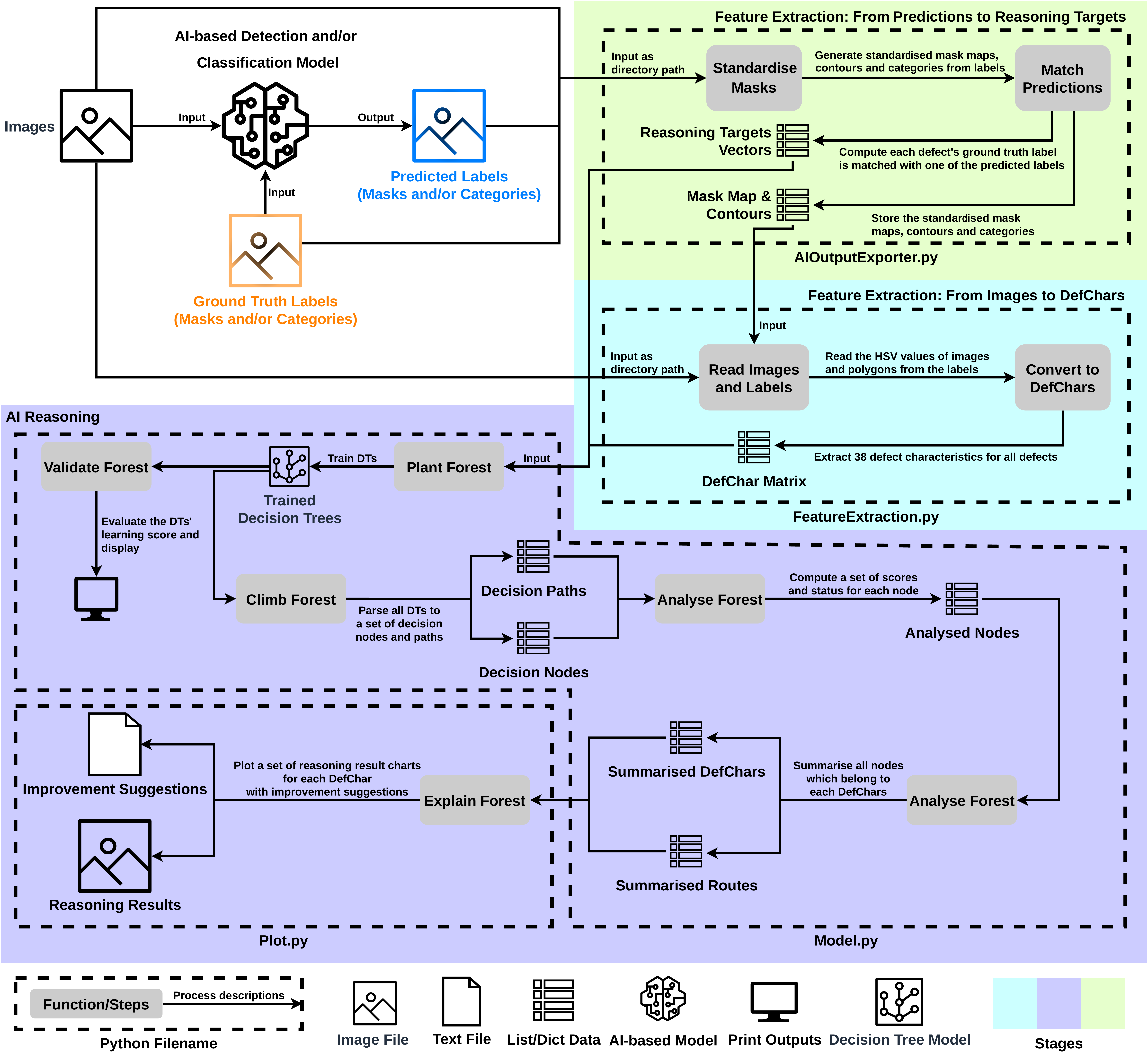}
    \caption{\gls{FM} toolkit structure overview.}
    \label{fig:arch_overview}
\end{figure*}

The \gls{FM} toolkit is a Python-based package library. Figure~\ref{fig:arch_overview} illustrates the architecture overview of the \gls{FM} toolkit. Zhang \etals \cite{zhangair} \gls{AI}-Reasoner serves as the foundation for the \gls{FM} toolkit, which is a post-hoc and model-agnostic framework with data explainability. As a result, the toolkit only requires input images, ground truth labels and predicted labels from the \gls{AI} model. The outputs of the toolkit are a set of charts and text-based improvement suggestions, which are stored as \textit{png} and \textit{txt} formatted files, respectively. The \gls{FM} toolkit comprises three stages to perform an \gls{AI} reasoning task:
\begin{itemize}
    \item Feature extraction from predictions to reasoning targets: This stage converts all predictions into a set of reasoning target vectors based on the \gls{AI} model's prediction tasks, such as detection, classification, or joint detection and classification. The reasoning targets include labels such as ``detected'', ``undetected'', ``correctly classified'', and ``misclassified''. Additionally, this stage standardises the mask-based labels to maintain consistent mask maps and contours of the defect regions.
    \item Feature extraction from images to \glspl{DC}: In this stage, all defect images are processed to generate a \gls{DC} matrix. This is achieved by reading and processing the \gls{HSV} values and mask maps of the images.
    \item \gls{AI} reasoning stage: This stage takes the \gls{DC} matrix and reasoning target vectors as input to analyse the importance of each \gls{DC} in causing correct or incorrect predictions by the \gls{AI} model. It involves six steps: \textit{plant forest}, \textit{validate forest}, \textit{climb forest}, \textit{analyse forest}, \textit{summarise forest}, and \textit{explain forest}. Finally, this stage generates improvement recommendations in the form of text and visualisations to explain the results and provide reasoning to users.
\end{itemize}

\section{Implementation}
\label{sec:impl}


The \gls{FM} toolkit is implemented using various packages, including \textit{scikit-learn}, \textit{OpenCV}, \textit{NumPy}, \textit{tqdm}, \textit{matplotlib}, \textit{shapely}, \textit{pillow}, and \textit{polygenerator}.

\begin{itemize}
    \item \textit{NumPy} for processing array-based and matrix-based data.
    \item \textit{OpenCV} and \textit{pillow:} image data processing, including computing the mask map and contours.
    \item \textit{Shapely} and \textit{polygenerator:} converting polygon contours into shape-based \glspl{DC}.
    \item \textit{scikit-learn:} \gls{DT} model implementation that is utilised for the \gls{AI} reasoning functionalities.
    \item \textit{matplotlib:} generating the reasoning result charts.
\end{itemize}

The hardware requirements for running the \gls{FM} toolkit include at least 8 gigabytes of RAM and any Intel/AMD CPU; a GPU is not required.

\subsection{Data Preparation}
\label{sec:dp}

\begin{figure}[!t]
    \centering    \includegraphics[width=0.7\linewidth]{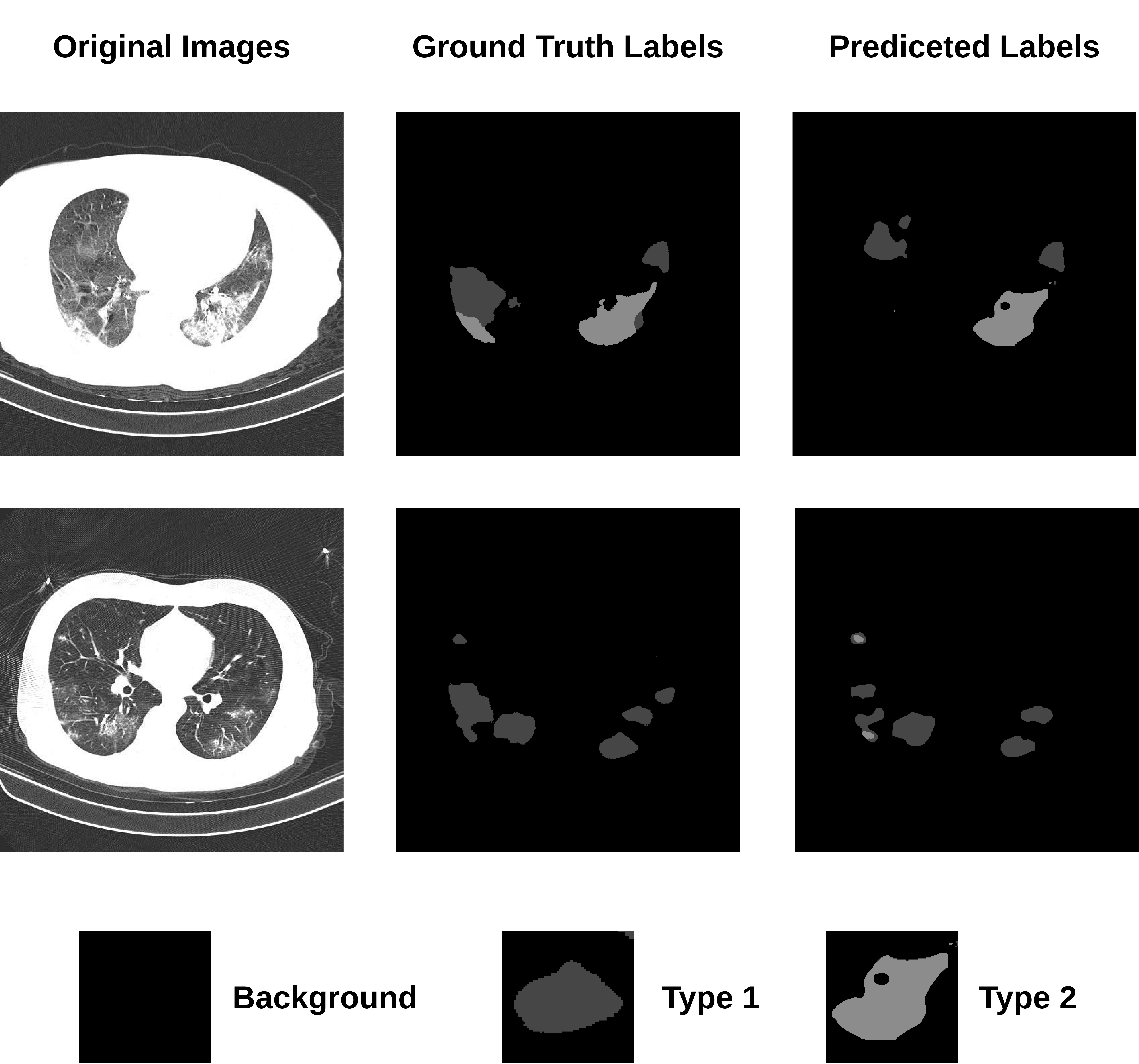}
    \caption{Example of input data for \gls{FM} toolkit, the real values in the highlighted area are the categories of defects $C \in \{1,2\}$.}
    \label{fig:example}
\end{figure}

A classic dataset usually consists of pre-processed images and their corresponding mask-based ground truth labels (stored in \textit{png} format) when applied in an \gls{AI} model. Once the detection and/or classification tasks are performed by the \gls{AI} model, its prediction results need to be converted into mask-based labels, mirroring the format of the ground truth labels. Figure~\ref{fig:example} provides an example of the input data structure. The images, ground truth labels and predicted labels should be stored in three separate folders.
\subsection{Feature Extraction: From Predictions to Reasoning Targets}
\begin{minted}[
bgcolor=LightGray,
fontsize=\small]
{python}
def process_dir(imgs_path,gt_masks_path,
predicted_masks_path,contain_type,only_type):
    ...
    return label_data={"mask map":[...],
           "contours":[...],
           "reasoning_targets":{...}}
\end{minted}

This stage involves the extraction of reasoning target vectors from \gls{AI} predictions, as well as the mask map and contours from the ground truth labels. This can be accomplished by calling the \textit{process$\_$dir()} function; the required inputs for this function are:
\begin{itemize}
    \item The path to the images folder.
    \item The path to the ground truth labels folder.
    \item The path to the predicted labels folder.
\end{itemize}
Additionally, two Boolean values, \textit{contained$\_$type} and \textit{only$\_$type}, are required to specify the extraction of reasoning target vectors for different tasks, such as a detection task (\textit{contained$\_$type = False}, \textit{only$\_$type = False}), a classification task (\textit{contained$\_$type = True}, \textit{only$\_$type = True}), or a joint detection and classification task (\textit{contained$\_$type = True}, \textit{only$\_$type = True}). The output of the function is a dictionary that contains the mask maps, contours, and reasoning target vectors for all ground truth defects.

\subsection{Feature Extraction: From Images to \Glspl{DC}}
\begin{minted}[
bgcolor=LightGray,
fontsize=\small]
{python}
def checkImage(imgs_path):
    ...
def readLabel(label_dict):
    ...
def loadData():
    ...
def featureExtract():
    ...
    return defchar={"defect_1":{...},
                    "defect_2":{...},...}
\end{minted}

This stage involves a sequence of four functions: \textit{checkImage()}, \textit{readLabel()}, \textit{loadData()}, and \textit{featureExtract()}. The \textit{checkImage()} function is responsible for loading the images, while the \textit{readLabel()} function loads the labels, which include mask maps and contours. The \textit{loadData()} function reads the images and labels and creates a dictionary with basic information about the images and labels. The \textit{featureExtract()} function is utilised to extract the \glspl{DC} and stores them in the dictionary. These functions are called in order to extract the \glspl{DC} from the images. 

\subsection{Pre-processing before AI Reasoning}
\begin{minted}[
bgcolor=LightGray,
fontsize=\small]
{python}
def convert2List(defchar,feature_list,
label_dict):
    ...
    return id_list=[defect_id,...],
           feature_list,
           data=[[defchar 1],[defchar 2],...],
           targets={"detected":[...],
                    "undetected":[...],...}
def load_feature_data(data):
    ...
def load_target_data(targets,target_name):
    ...
\end{minted}
In this section, there are three utility functions (i.e.\ \textit{convert2List()}, \textit{load$\_$feature$\_$data()}, and \textit{load$\_$target$\_$data()}) implemented in the \textit{Model.py} file. The \textit{convert2List()} function returns several array-based lists derived from the feature extraction stage, including an id list, a feature list, a \glspl{DC} matrix and reasoning target vectors. The \textit{load$\_$feature$\_$data()}, and \textit{load$\_$target$\_$data()} functions are utilised to correspondingly load and prepare the feature and reasoning target data for the \gls{AI}-Reasoner model. These functions are responsible for converting the dictionary-based data into array-based data and loading it into the \gls{AI}-Reasoner model. 

\subsection{Plant Forest}
\begin{minted}[
bgcolor=LightGray,
fontsize=\small]
{python}
def plant_forest(n_tree=200):
    ...
    return model=[DT_1,DT_2,...]
\end{minted}

The \textit{plant$\_$forest()} function is responsible for building and training multiple \gls{DT} models using the loaded data. The function allows for an optional input to specify the desired number of \gls{DT} models, which determines the number of decision rules generated. By invoking the \textit{plant$\_$forest()} function, the \gls{DT} models are constructed and trained using the loaded data. 
\subsection{Validate Forest}
\begin{minted}[
bgcolor=LightGray,
fontsize=\small]
{python}
def val_forest(model, feature_list):
    ...
    return good_learned=bool(),
           eval_=[learning_score, TPR, TNR],
           error_feature=[...]
\end{minted}

The \textit{val$\_$forest()} function is responsible for evaluating the learning capability of the \gls{AI}-Reasoner. It computes the overall learning scores, including \gls{TPR} and \gls{TNR}, by averaging the learning scores of each trained \gls{DT} model. Additionally, the function provides a set of \glspl{DC} where these \gls{DC} made mistakes during the reasoning task. 

\subsection{Climb Forest}
\begin{minted}[
bgcolor=LightGray,
fontsize=\small
]
{python}
def climb_forest(model,feature_list):
    ...
    return path=[...],node=[...],route=[...]
\end{minted}
The \textit{climb$\_$forest()} function is responsible for parsing all decision nodes and paths from the trained \gls{DT} models. It stores the parsed information in three lists: decision paths, routes, and decision nodes. Additionally, the list of \glspl{DC} names is required for the parsing process; the list can be accessed by calling \textit{FeatureExtraction.feature$\_$list}.
\subsection{Analyse Forest}
\begin{minted}[
bgcolor=LightGray,
fontsize=\small
]
{python}
def analyse_forest(path,node,error):
    ...
    return analysed_node=[...]
\end{minted}

The \textit{analyse$\_$forest()} function takes the parsed decision nodes and paths as input. It then computes a set of values for each node, which helps analyse the importance of the node in reasoning the \gls{AI} predictions. The function outputs a list that extends the input node list with the computed values; this extended list provides additional information and insights about each node, allowing for a more in-depth analysis of the reasoning process.
\subsection{Summarise Forest}
\begin{minted}[
bgcolor=LightGray,
fontsize=\small
]
{python}
def summary_forest(analysed_node,feature_list,
route,feature_range):
    ...
    return summary={"defchar 1":{...},...},
           route_to_1=[...],
           route_to_0=[...]
\end{minted}

The \textit{summary$\_$forest()} function computes the importance for each \gls{DC} to reflect the importance of each \gls{DC} in influencing the \gls{AI} predictions. It takes several inputs, including the analysed nodes list, parsed route, \glspl{DC} list, and value ranges. The function outputs a dictionary that contains the computed scores for each \gls{DC} to provide a comprehensive overview of the importance of each \gls{DC}.

\subsection{Explain Forest}
\begin{minted}[
bgcolor=LightGray,
fontsize=\small
]
{python}
def explain_forest(report,feature_list,
save_path,route_plot=None):
    ...
\end{minted}

The \textit{explain$\_$forest()} function is responsible for generating a set of charts and providing improvement suggestions based on the summarised overviews from the previous stage. The input includes the generated dictionary from \textit{summary$\_$forest()} function, \glspl{DC} list and folder path where the reasoning results will be saved. Furthermore, a set of important decision routes can be optionally plotted by setting \textit{route$\_$plot} as true. The reasoning result charts are saved in the \textit{png} format, while the improvement suggestions are stored in a \textit{txt} format file.

\section{Performance}
\label{sec:perf}

\begin{table*}[ht]
\centering
\caption{Information of four datasets; where N/A represents no such tasks.}
\begin{tabular}{lcccccc}
\toprule
\multirow{2}{*}{Dataset} & \multirow{2}{*}{Image Size} & \multicolumn{4}{c}{Number of Defects for Each Reasoning Target (percentage)} & \multirow{2}{*}{Total} \\ \cmidrule{3-6}
& & detected & undetected  & correctly classified  & misclassified & \\\midrule
Chest CT & 512$\times$512                       & 244 & 147 & 215 & 29 & 391\\
Heatsink Defect & 256$\times$256                & 804 & 368 & 774 & 30 & 1172\\
Kvasir-SEG & 256$\times$256                     & 104 & 3 & N/A & N/A & 107\\
Wind Turbine Blade Defect & 1920$\times$1080    & 311 & 55 & 296 & 15 & 366\\
\bottomrule
\end{tabular}

\label{tab:data_dist}
\end{table*}

In this section, the \gls{FM} toolkit was applied to four different \gls{AI}-based defect detection and classification models using different datasets. The models include a \textit{COVID-CT-mask-net} trained on a chest CT image dataset \cite{ccmn}, a \textit{lightweight fully convolutional network} trained on a heatsink defect image dataset \cite{lfcn}, a \textit{deep residual U-Net++} trained on the Kvasir-SEG dataset \cite{run}, and an \textit{image-enhanced mask R-CNN} trained on a wind turbine blade defect image dataset \cite{IEMRCNN}. Table~\ref{tab:data_dist} presents the defect distributions of these four datasets, providing an overview of the reasoning targets and quantities of defects present in each dataset. Additionally, the running time in each step of the \gls{FM} toolkit was recorded to analyse its performance. This allows for an assessment of the toolkit's efficiency and provides insights into potential areas for improvement. A computer with an AMD 16-core CPU and 32 gigabytes RAM was used to evaluate performance in execution time.

\begin{figure}[!t]
    \centering
\includegraphics[width=\linewidth]{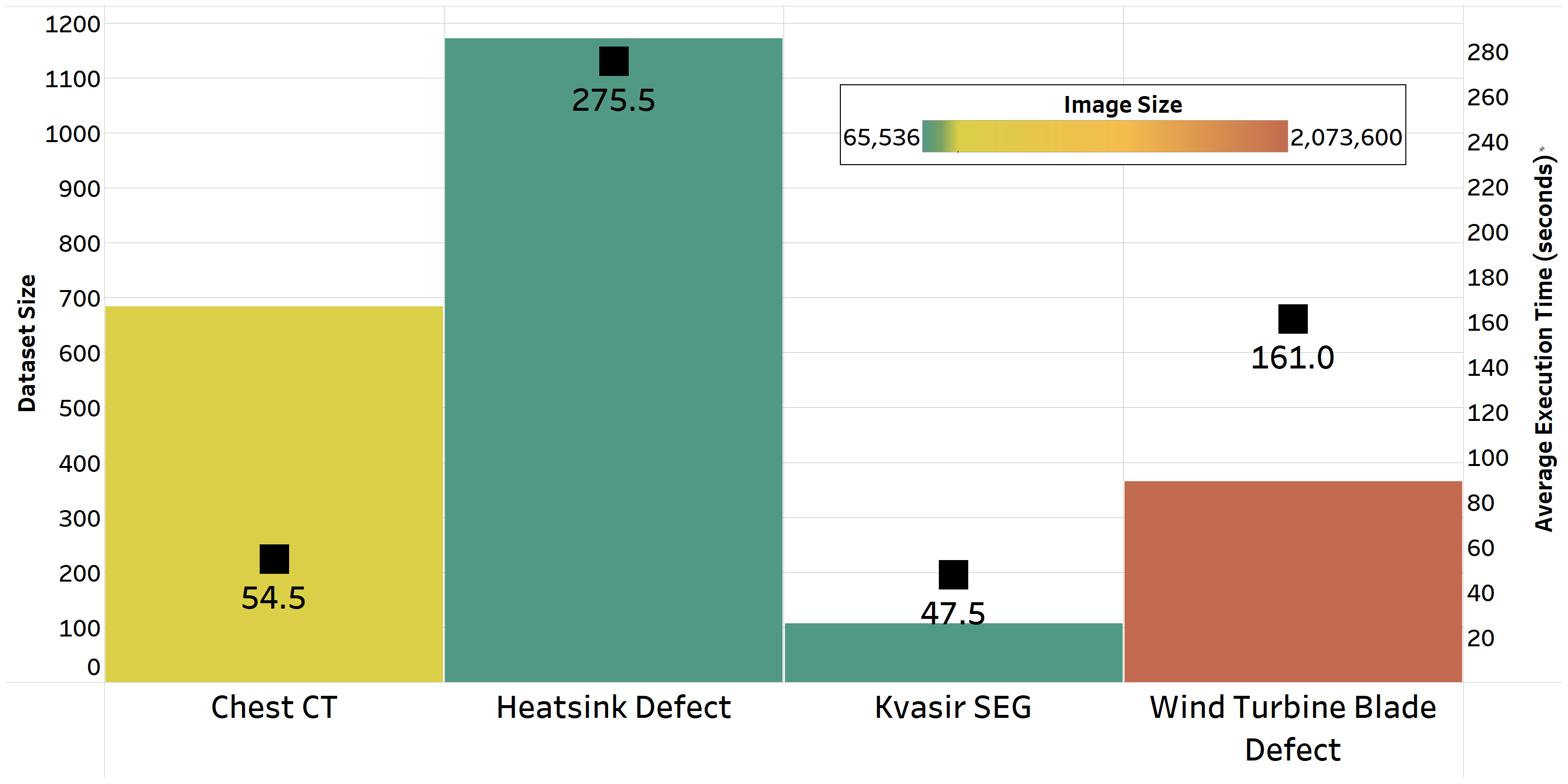}
    \caption{Average execution time of the FM toolkit for different datasets. The black dots represent the average execution time, indicated on the right axis. The bars represent the dataset size, indicated on the left axis, and the colours of the bars correspond to the dataset's image size, as shown in the legends.}

    \label{fig:time2}
\end{figure}
\begin{figure*}[!t]
    \centering
\includegraphics[width=0.75\linewidth]{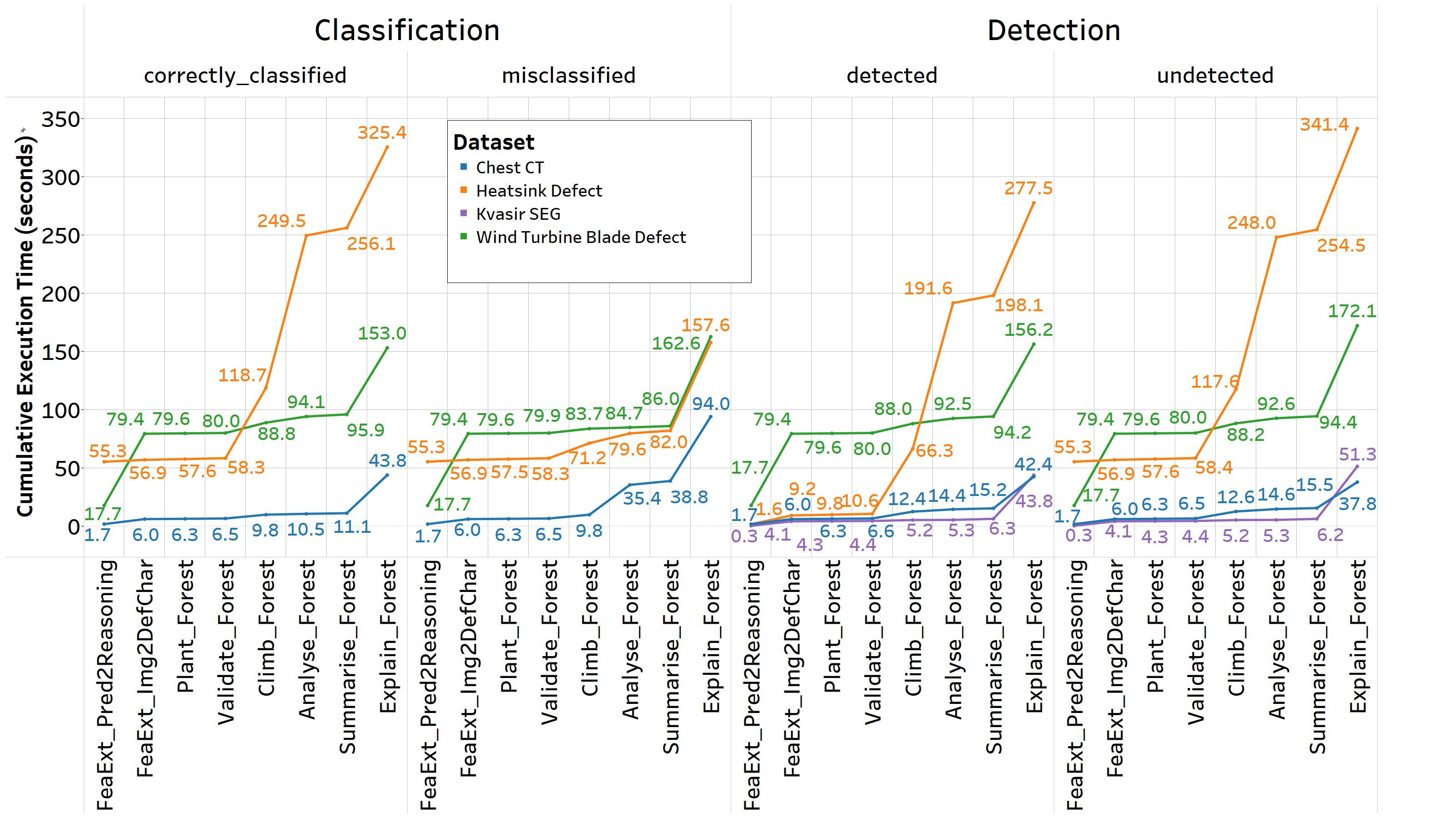}
    \caption{Cumulative execution time of each stage in the \gls{FM} toolkit for reasoning targets of different datasets.}
    
    \label{fig:time1}
\end{figure*}

Figure~\ref{fig:time2} illustrates the execution times when applying the \gls{FM} toolkit to different datasets. It shows that the execution time is positively correlated with the dataset size and image size, indicating that larger datasets or images require longer execution times. Figure~\ref{fig:time1} illustrates the cumulative execution times of each stage for different reasoning targets. Among the various stages of the \gls{FM} toolkit, \textit{feature extraction from images to \glspl{DC}}, \textit{climb forest}, \textit{analyse forest}, and \textit{explain forest} consumed relatively more time compared to other stages. In summary, the \gls{FM} toolkit can complete a reasoning task in at least 40 seconds. However, the execution time may increase when dealing with datasets containing large amounts or large-size images.

\section{Tutorial}
\label{sec:tuto}
This section provides a tutorial for adopting the \gls{FM} toolkit to reason with the outputs of an \gls{AI} model.

\noindent\textbf{Step 1:} Import the \gls{FM} toolkit.
\begin{minted}[
bgcolor=LightGray,
fontsize=\small
]
{python}
from ForestMonkey import AIOutputExporter
from ForestMonkey import FeatureExtraction
from ForestMonkey import Model
from ForestMonkey import Plot
\end{minted}
\noindent\textbf{Step 2:} Set the directory paths of original images, ground truth labels and predicted labels from the \gls{AI} model; the related data preparation is described in Section~\ref{sec:dp}.
\begin{minted}[
bgcolor=LightGray,
fontsize=\small]
{python}
images="/path/.../"
gt_labels="/path/.../"
predicted_labels="/path/.../"
\end{minted}
\noindent\textbf{Step 3:} Feature extraction: From predictions to reasoning targets
\begin{minted}[
bgcolor=LightGray,
fontsize=\small,
,
]
{python}
contain_type=True
type_only=False
label_data=AIOutputExporter.process_bydir(
           images,gt_labels,predicted_labels)
\end{minted}
\noindent\textbf{Step 4:} Feature extraction: From images to \glspl{DC}
\begin{minted}[
bgcolor=LightGray,
fontsize=\small,
,
]
{python}
FeatureExtraction.checkImages(images)
FeatureExtraction.readLabel(label_data)
FeatureExtraction.loadData()
defchar=FeatureExtraction.featureExtract()
\end{minted}
\noindent\textbf{Step 5:} Pre-processing before AI reasoning 
\begin{minted}[
bgcolor=LightGray,
fontsize=\small
]
{python}
feature_list = FeatureExtraction.feature_list
id_list,feature,data,target=Model.convert2List
(defchar,feature_list,label_data)
Model.load_feature_data(data)
\end{minted}
\noindent\textbf{Step 6:} Execute \gls{AI} reasoning task for all reasoning targets
\begin{minted}[
bgcolor=LightGray,
fontsize=\small
]
{python}
for t in target.keys():
    Model.load_target_data(target[t],t)
    model = Model.plant_forest()
    tree_validated,scores,errors=Model.
val_forest(model,feature)
    path,node,route=Model.climb_forest(model,
feature_name=feature)
    analysed_node=Model.analyse_forest(path,
node,error)
    report,route_t,route_nt=Model.
summary_forest(analysed_node,route,feature,
FeatureExtraction.get_FeatureRange())
    Plot.explain_forest(report,feature,
"/directory/path/to/save/"+t)
\end{minted}

\begin{figure}[!t]
    \centering
    \includegraphics[width=\linewidth]{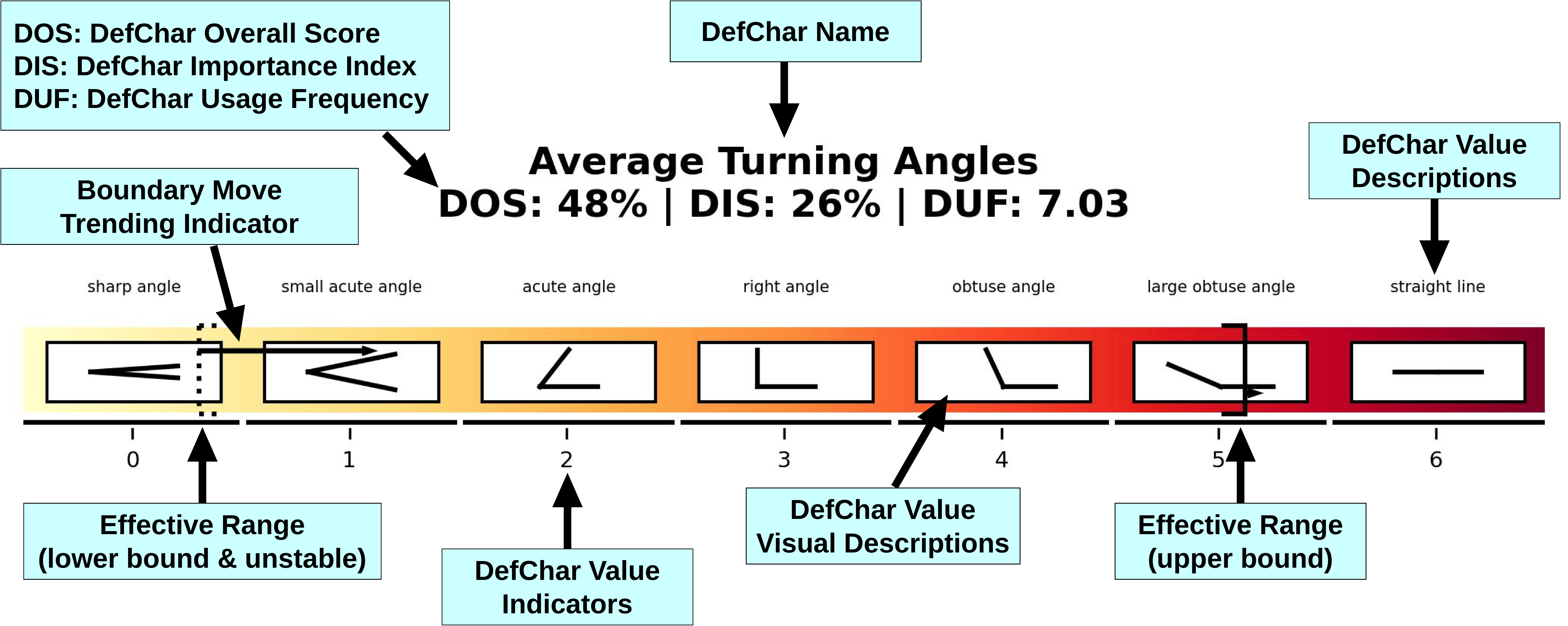}
    \caption{An example chart illustrating one of the \gls{DC}'s reasoning results generated by the \gls{FM} toolkit; the blue textual boxes represent descriptions of each element in the chart.}
    \label{fig:example_output}
\end{figure}

After completing these six steps, the reasoning results will be saved in the specified directory. Figure~\ref{fig:example_output} illustrates an example chart generated by the FM toolkit. The chart allows users to understand the significance of each \gls{DC} and its value range in influencing the AI model's correct or incorrect predictions in detection and/or classification tasks. Additionally, users can refer to the \textit{improvement$\_$recommendations.txt} file for improvement suggestions on how to enhance their dataset and model based on the reasoning results.

\section{Conclusion}
\label{sec:conclusion}

This paper presents the integration of Zhang et al.'s \cite{zhangair} AI-Reasoner framework into a toolkit called \gls{FM}, implemented in Python. The \gls{FM} toolkit can be easily used by importing it as a Python package, and a detailed tutorial is provided to guide users in utilising the toolkit effectively. Furthermore, the FM toolkit is evaluated by applying it to four different AI-based models with diverse datasets to assess its execution performance. 
In terms of future work, several enhancements are suggested for the \gls{FM} toolkit. Firstly, the implementation of GPU-enabled parallel computations could be explored to accelerate the execution speed. Additionally, the development of interactive interfaces would enhance user experience and make the toolkit more user-friendly. Furthermore, visualisations of the \glspl{DC} could be incorporated to provide users with a better understanding of the reasoning process.


\bibliographystyle{IEEEtran}
\bibliography{IEEEfull,root}                                                              

\end{document}